\documentclass[10pt,twocolumn,letterpaper]{article}

\usepackage[pagenumbers]{cvpr} 

\usepackage{amsmath,amsfonts,bm}

\def\Figref#1{Figure~\ref{#1}}

\def\Secref#1{Section~\ref{#1}}

\def\eqref#1{equation~\ref{#1}}

\def\1{\bm{1}}

\def\vd{{\bm{d}}}

\def\vf{{\bm{f}}}

\def\vr{{\bm{r}}}
\def\vs{{\bm{s}}}
\def\vt{{\bm{t}}}

\def\vx{{\bm{x}}}
\def\vy{{\bm{y}}}

\def\mC{{\bm{C}}}

\def\mE{{\bm{E}}}
\def\mF{{\bm{F}}}
\def\mG{{\bm{G}}}

\def\mI{{\bm{I}}}

\def\mK{{\bm{K}}}

\def\mM{{\bm{M}}}

\def\mO{{\bm{O}}}
\def\mP{{\bm{P}}}
\def\mQ{{\bm{Q}}}
\def\mR{{\bm{R}}}

\def\mW{{\bm{W}}}
\def\mX{{\bm{X}}}
\def\mY{{\bm{Y}}}

\DeclareMathAlphabet{\mathsfit}{\encodingdefault}{\sfdefault}{m}{sl}
\SetMathAlphabet{\mathsfit}{bold}{\encodingdefault}{\sfdefault}{bx}{n}

\def\sS{{\mathbb{S}}}

\newcommand{\Ls}{\mathcal{L}}
\newcommand{\R}{\mathbb{R}}

\newcommand{\normlone}{L^1}
\newcommand{\normltwo}{L^2}

\usepackage{pifont}
\usepackage{multirow}
\usepackage{booktabs}       
\usepackage[table,xcdraw]{xcolor}
\newcommand{\parhead}[1]{\noindent \textbf{{#1}}}

\newcommand{\ours}{ComPose}
\newcommand{\so}{\mathrm{SO(3)}}

\newcommand{\Softmax}{\phi}

\newcommand{\iou}[1]{$\text{IoU}_{#1}$}
\newcommand{\ducm}[2]{$#1^{\circ}#2\text{cm}$}

\newcommand{\img}{\mI^{\mathrm{rgb}}}
\newcommand{\pts}{\mP^{\mathrm{part}}}
\newcommand{\kpt}{\mP^{\mathrm{kpt}}}
\newcommand{\com}{\mP^{\mathrm{com}}}
\newcommand{\fold}{\mP^{\mathrm{fold}}}
\newcommand{\cad}{\mM^{\mathrm{cad}}}
\newcommand{\cadObs}{\mM^{\mathrm{obs}}}
\newcommand{\ptsSup}{\mP^{\mathrm{sup}}}
\newcommand{\ptsFeat}{\mF^{\mathrm{part}}}
\newcommand{\kptQuery}{\mQ^{\mathrm{kpt}}}
\newcommand{\kptFeat}{\mF^{\mathrm{kpt}}}
\newcommand{\ptsNum}{N^{\mathrm{part}}}

\newcommand{\kptNum}{N^{\mathrm{kpt}}}
\newcommand{\kptInd}{n}
\newcommand{\kptIndSecond}{m}
\newcommand{\comNum}{N^{\mathrm{com}}}

\newcommand{\foldNum}{N^{\mathrm{fold}}}

\newcommand{\coor}{\mC}
\newcommand{\coorNum}{N}
\newcommand{\feat}{\mF}
\newcommand{\featDim}{D}

\newcommand{\knn}{\mP^{\mathrm{knn}}}
\newcommand{\knnFeat}{\mF^{\mathrm{knn}}}
\newcommand{\geoFeat}{\mF^{\mathrm{geo}}}
\newcommand{\geoLocal}{\mE^{\mathrm{l}}}
\newcommand{\geoGlobal}{\mE^{\mathrm{g}}}
\newcommand{\knnNum}{N^{\mathrm{knn}}}

\newcommand{\textPn}{\mathrm{pn}}
\newcommand{\textDino}{\mathrm{dino}}
\newcommand{\textInit}{\mathrm{init}}
\newcommand{\textCom}{\mathrm{com}}
\newcommand{\textVis}{\mathrm{vis}}
\newcommand{\textMiss}{\mathrm{miss}}

\newcommand{\textKpt}{\mathrm{kpt}}
\newcommand{\textGeo}{\mathrm{geo}}
\newcommand{\textScore}{\mathrm{score}}
\newcommand{\textCand}{\mathrm{cand}}
\newcommand{\textAll}{\mathrm{all}}

\newcommand{\textGlobal}{\mathrm{global}}

\newcommand{\textGt}{\mathrm{gt}}

\newcommand{\textNocs}{\mathrm{nocs}}
\newcommand{\textCorr}{\mathrm{corr}}

\newcommand{\nocs}{\mO}
\newcommand{\relation}{\mG}
\newcommand{\score}{\vr}
\newcommand{\dis}{\vd}
\newcommand{\temperature}{\tau}
\newcommand{\comSet}{\sS}

\newcommand{\rotation}{\mR}
\newcommand{\translation}{\vt}
\newcommand{\size}{\vs}

\definecolor{cvprblue}{rgb}{0.21,0.49,0.74}
\usepackage[pagebackref,breaklinks,colorlinks,allcolors=cvprblue]{hyperref}

\def\paperID{****} 
\def\confName{CVPR}
\def\confYear{2026}

\title{ComPose: A Unified Completion-Pose Framework for Robust\\Category-Level Object Pose Estimation}

\author{Huan Ren$^{1,2}$ \enskip Yihan Chen$^{1,2}$ \enskip Chuxin Wang$^{1,2}$ \enskip Nailong Liu$^{3}$ \enskip Wenfei Yang$^{1,2}$\thanks{Corresponding author. Website: \href{https://renhuan1999.github.io/ComPose}{renhuan1999.github.io/ComPose}.} \enskip Tianzhu Zhang$^{1,2}$\\
\normalsize $^{1}$University of Science and Technology of China\\
\normalsize $^{2}$National Key Laboratory of Deep Space Exploration, Deep Space Exploration Laboratory\\
\normalsize $^{3}$Beijing Institute of Control Engineering\\
\texttt{\small \{rh\_hr\_666,yihanchen,wcx0602\}@mail.ustc.edu.cn \quad \{yangwf,tzzhang\}@ustc.edu.cn}
}

\begin{document}
\maketitle
\begin{abstract}
Category-level object pose estimation aims to predict the pose and size of arbitrary objects in specific categories. Existing methods struggle with the inherent incompleteness of observed point clouds, which limits their ability to capture complete object shapes for robust pose reasoning. While point cloud completion offers a promising solution, naively treating it as a separate preprocessing step for partial observations introduces compounding errors and additional computational overhead, ultimately hindering both accuracy and efficiency.
To address these challenges, we propose \ours{}, a novel unified framework that tightly integrates shape completion to provide complete geometric cues for enhanced pose estimation. At the core of \ours{} is a keypoint-based progressive completion module, which recovers full shape representations by progressively predicting a sparse set of keypoints and their surrounding dense point sets, empowering the keypoints to capture holistic object geometries. A geometric relation encoding module further enriches keypoint features with both local and global geometric context. In addition, we introduce a novel geometric relation consistency loss to enforce structural alignment between observed keypoints and their predicted NOCS coordinates, ensuring globally coherent coordinate transformations.
Extensive experiments on standard benchmarks demonstrate that our method outperforms state-of-the-art approaches without relying on category-level shape priors.
\end{abstract}    
\vspace{-2em}
\section{Introduction}
\label{sec:intro}

Category-level object pose estimation aims to predict the 6D pose and 3D size of arbitrary objects within predefined categories. As a fundamental task in 3D computer vision, it has attracted significant attention from the research community due to its extensive potential applications in fields such as robotic manipulation~\cite{robot1}, augmented reality~\cite{ar2}, and autonomous driving~\cite{drive1}. In contrast to traditional instance-level approaches~\cite{cvpr2019pvnet,cvpr2019densefusion,cvpr2021gdrnet}, category-level methods do not require instance-specific CAD models for inference, exhibiting stronger generalization in real-world scenarios.

\begin{figure}[t]
\centering
\includegraphics[width=1\linewidth]{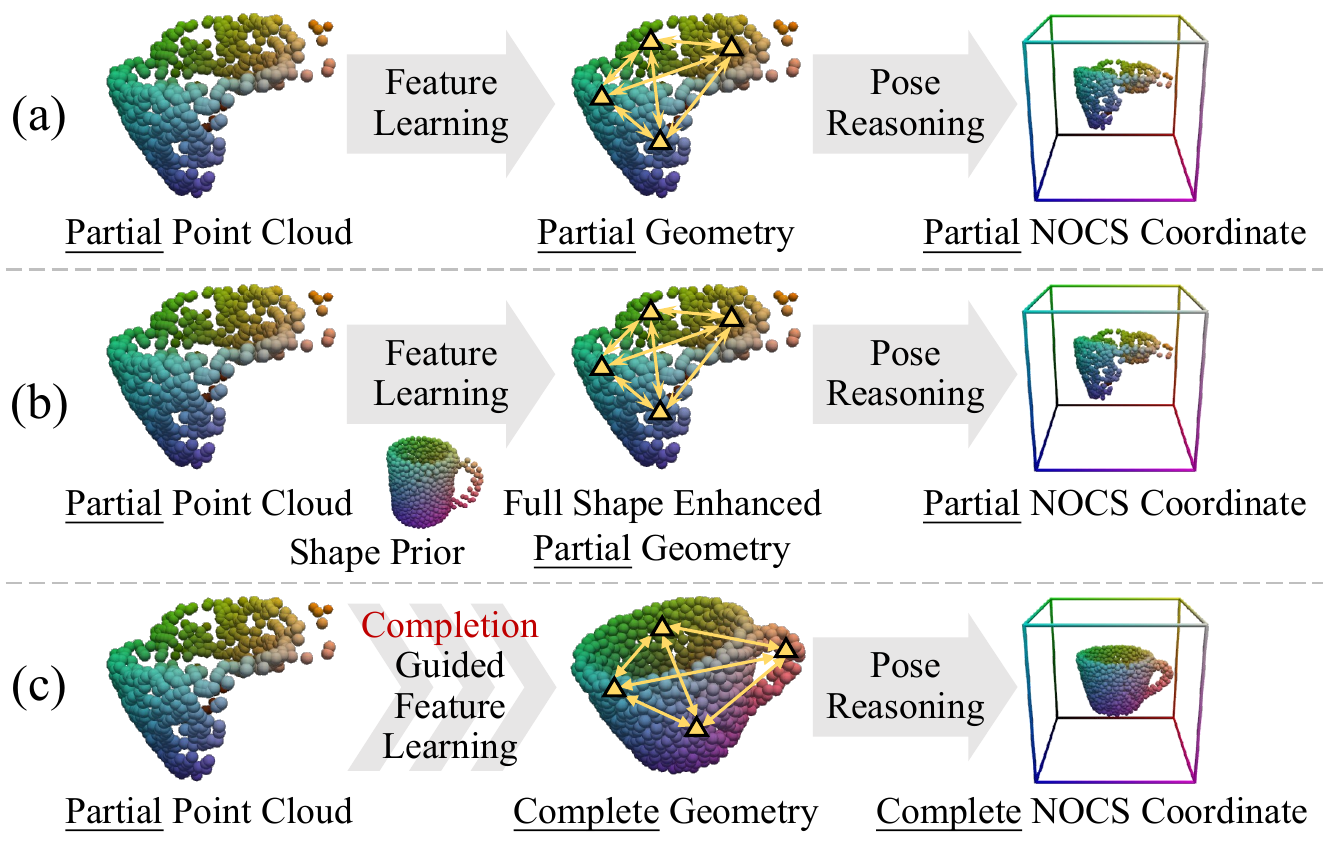}
\vspace{-1.75em}
\caption{%
Comparison of geometric representation strategies in category-level object pose estimation. (a) Classic methods directly encode geometric features from partial point clouds, which limits their ability to capture complete object structures. (b) Prior-based approaches resort to category-level shape priors~\cite{eccv2020spd} to enhance feature understanding of full object shapes, yet they still operate on incomplete geometries. (c) Our method explicitly integrates shape completion to recover complete geometries, facilitating more comprehensive and robust pose reasoning.
}
\vspace{-1.25em}
\label{fig:motivation_geometry}
\end{figure}

Existing category-level methods typically commence by extracting features from partial observations to comprehend object shapes, which subsequently guides either direct pose regression or the prediction of Normalized Object Coordinate Space (NOCS)~\cite{cvpr2019nocs} as an intermediate representation for pose fitting. Despite the considerable progress achieved by these approaches, their performance remains fundamentally constrained by the inherent incompleteness of partially observed point clouds. Specifically, depth cameras fail to capture the occluded backside of objects due to self-occlusion, yielding incomplete point clouds after back-projection.
As illustrated in \Figref{fig:motivation_geometry}(a), most previous methods~\cite{cvpr2022gpvpose,cvpr2024agpose,cvpr2025spotpose,iccv2025cleanpose} encode geometric structures directly from such partial point clouds, which restricts their ability to capture complete object shapes for robust pose reasoning. To alleviate this limitation, several works~\cite{eccv2022dpdn,cvpr2025gcepose} incorporate category-level shape priors~\cite{eccv2020spd} to enhance the comprehension of full shape context at the feature level, as shown in \Figref{fig:motivation_geometry}(b). Nevertheless, these methods only provide indirect shape cues \textit{in the canonical space} while still operate on intrinsically incomplete shape representations \textit{in the observation space}, leaving the fundamental issue of geometric incompleteness unresolved. Moreover, acquiring shape priors requires collecting extensive CAD models and training an extra autoencoder, which is labor-intensive and costly.

\begin{figure}[t]
\centering
\includegraphics[width=1\linewidth]{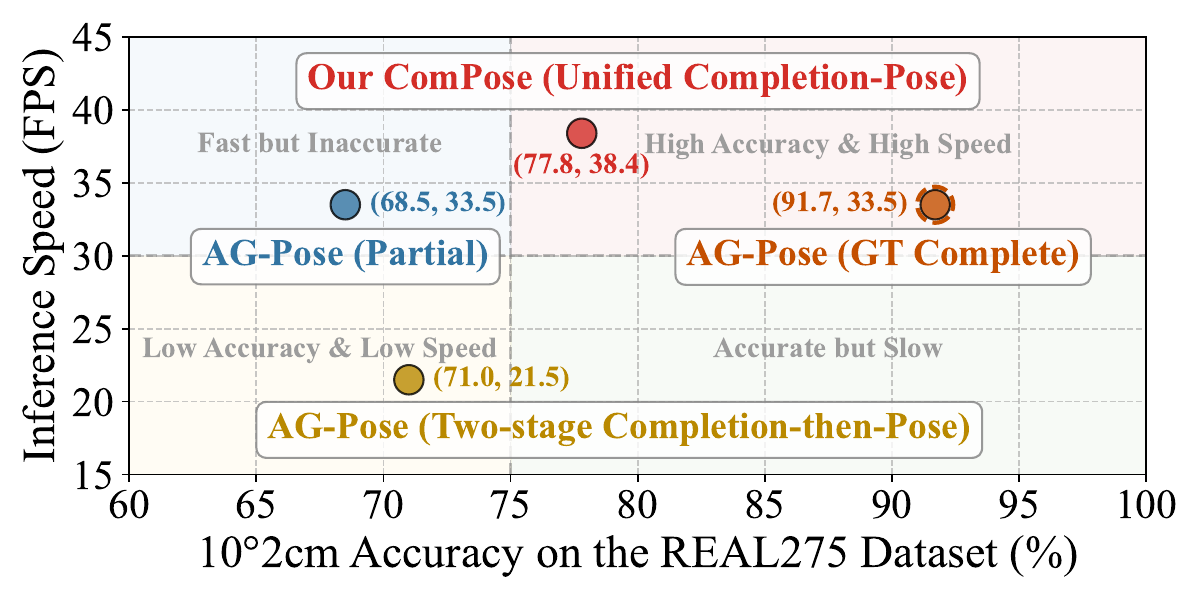}
\vspace{-2em}
\caption{%
Accuracy and inference speed comparison for the depth-only versions of different methods. The dashed circle indicates the performance upper bound achieved using ground-truth complete point clouds as input. Our ComPose achieves the best balance between accuracy and efficiency with 38.4 FPS on an RTX3090Ti GPU. More implementation details are provided in \Cref{sec:ablation}.
}
\vspace{-1em}
\label{fig:motivation_accuracy_fps}
\end{figure}

To address these challenges, we draw inspiration from recent advancements in point cloud completion~\cite{iccv2021pointr,tpami2023adapointr} and explore reconstructing complete object shapes directly in the observation space. As illustrated in \Figref{fig:motivation_geometry}(c), the completed point clouds provide a more comprehensive geometric representation of objects, which is crucial for robust pose reasoning. To quantitatively assess the merit of complete geometries, we conduct an oracle experiment by replacing the \textit{partial point cloud} inputs in the depth-only version of the leading AG-Pose~\cite{cvpr2024agpose} network with \textit{ground-truth complete point clouds}, while keeping the network architecture unchanged.
As shown in \Figref{fig:motivation_accuracy_fps}, the \ducm{10}{2} accuracy improves dramatically from 68.5\% to 91.7\%, highlighting the significant upper-bound performance gain enabled by full shape information.
This finding naturally suggests a straightforward solution that employs a point cloud completion network to first recover the full shape and then feeds it into a standard pose estimator, such as AG-Pose. However, such a naive two-stage pipeline suffers from compounding errors and introduces additional computational overhead, which compromise both accuracy and efficiency. As shown in \Figref{fig:motivation_accuracy_fps}, even an end-to-end jointly optimized variant yields only a marginal improvement of the \ducm{10}{2} accuracy to 71.0\% while reducing inference speed from 33.5 FPS to 21.5 FPS. This indicates that simply cascading completion and pose estimation networks falls short of fully exploiting the potential of shape completion and incurs a notable efficiency trade-off.
These observations thus raise a pivotal question: \textit{how can we effectively and efficiently integrate the complete geometric cues recovered from point cloud completion to enhance object pose estimation?}

Motivated by the above discussions, we propose \textbf{\ours{}}, a novel category-level framework that seamlessly unifies point cloud \textbf{com}pletion and object \textbf{pose} estimation within a single network.
Unlike naive cascaded pipelines that treat completion as a separate preprocessing step, \ours{} tightly integrates completion as a task-driven internal component, thereby enhancing the comprehension of complete object shapes in a more effective and efficient manner. At the core of our framework is a keypoint-based progressive completion module that reconstructs full object shapes from partial observations by predicting a sparse set of complete keypoints along with their surrounding dense point sets. This design not only yields comprehensive shape representations in the observation space, but also empowers keypoints to capture complete geometries from a holistic perspective. Building on this, we incorporate a geometric relation encoding module to enrich keypoint features with both local and global geometric context, which facilitates keypoint-wise prediction of NOCS coordinates. To further ensure globally coherent coordinate transformations, we introduce a geometric relation consistency loss grounded in relational modeling~\cite{cvpr2023pmil,tpami2025crc}, which enforces structural alignment between observed keypoints and their predicted NOCS counterparts. In contrast to conventional point-to-point coordinate mapping losses, the proposed relation-based constraint captures higher-order structural cues, resulting in more robust and precise object pose estimation.

In summary, the contributions of this work are fourfold:
(1) We pioneer a novel paradigm that leverages the complete geometric cues recovered from point cloud completion to enhance the understanding of holistic object shapes, unlocking substantial potential for advanced pose reasoning.
(2) We propose a unified framework that seamlessly integrates shape completion and pose estimation into a single network, delivering an effective and efficient solution for category-level object pose estimation.
(3) We introduce a carefully designed approach that acquires complete shape representations through keypoint-based progressive completion, and further incorporates geometric relation encoding and consistency constraints for robust coordinate transformations.
(4) Extensive experimental results demonstrate the superior performance of our method over state-of-the-art approaches. Notably, our depth-only model achieves a significant 9.1\% improvement on the \ducm{10}{2} metric of the REAL275 dataset without relying on shape priors. 

\section{Related Works}
\label{sec:related}

\subsection{Category-level Object Pose Estimation}

Category-level object pose estimation aims to determine the 6D pose and 3D size of novel objects in specific categories, without requiring instance-specific CAD models at test time. A core challenge in this field lies in understanding object shapes with significant intra-class variations for robust pose reasoning, especially from partial and incomplete observations.
One class of works~\cite{cvpr2021fsnet,iccv2021dualposenet} directly regresses object poses from point cloud data by extracting discriminative geometric features, often incorporating geometry-guided constraints~\cite{cvpr2022gpvpose} or hybrid-scope geometric perception~\cite{cvpr2023hspose}. Another line of methods~\cite{iccv2023istnet,cvpr2025spotpose} leverages Normalized Object Coordinate Space (NOCS)~\cite{cvpr2019nocs} as a shared canonical representation to align various object instances, which enables the establishment of dense point-wise~\cite{eccv2022dpdn} or sparse keypoint-wise~\cite{cvpr2024agpose} correspondences between camera and object coordinate spaces for subsequent pose fitting. However, both types of methods are inherently constrained by the incomplete nature of partial point clouds, which limits their ability to capture complete object shapes.
To alleviate this, prior-based methods~\cite{iccv2021sgpa,cvpr2025gcepose} incorporate category-level shape priors~\cite{eccv2020spd} to enhance the feature-level perception of full shape context. For instance, SPD~\cite{eccv2020spd} predicts deformation fields from shape priors to reconstruct 3D object models in the canonical space, which in turn guides the prediction of NOCS coordinates from partial point clouds. 
Nevertheless, these approaches still operate on incomplete shape representations in the observation space, leaving the fundamental issue of geometric incompleteness unresolved.
In contrast, our \ours{} explicitly integrates shape completion to recover complete shape representations directly within the observation space, enabling the network to capture holistic object geometries and facilitating more robust pose reasoning without relying on external shape priors.

\subsection{Point Cloud Completion}

Point cloud completion aims to reconstruct the complete 3D shape from partial point clouds, a naturally arising problem in real-world scenarios where sensor data is often incomplete due to inevitable object self-occlusion. Early methods typically adopt an encoder-decoder architecture, where a global shape representation is extracted from the visible part and decoded into the complete point cloud. For instance, FoldingNet~\cite{cvpr2018foldingnet} proposes a folding-based decoder that deforms a 2D grid onto the 3D surface, while PCN~\cite{3dv2018pcn} introduces a coarse-to-fine decoding strategy to progressively recover fine-grained geometry.
However, these approaches primarily focus on global shape embeddings, which limits their capacity to explicitly model the interactions between visible and missing regions. To deal with this limitation, recent approaches such as PoinTr~\cite{iccv2021pointr} reformulate point cloud completion as a set-to-set translation problem, leveraging Transformer-based attention mechanisms~\cite{attention} to model long-range dependencies. AdaPoinTr~\cite{tpami2023adapointr} further improves upon this by introducing an adaptive query generation strategy and an auxiliary denoising task to boost completion quality.
Despite these advances, most methods treat point cloud completion as a standalone task, with limited exploration of its integration into downstream applications such as category-level object pose estimation.
A notable attempt is DR-Pose~\cite{iros2023drpose}, which leverages an off-the-shelf completion network~\cite{iccv2021pointr} to recover the missing object parts. However, the completed shapes are merely used to guide the deformation of shape priors as in SPD~\cite{eccv2020spd} and remain decoupled from the actual pose reasoning process. In contrast, we tightly integrate completion as an internal component of our unified \ours{} framework to enhance the understanding of complete object shapes for robust pose reasoning.

\section{Method}
\label{sec:method}

\parhead{Task Definition.}
Given an RGB-D image, an off-the-shelf Mask R-CNN~\cite{maskrcnn} network is first used to obtain instance masks, yielding the cropped RGB image $\img \in \R^{H \times W \times 3}$ and segmented depth image for each instance. The partial point cloud $\pts \in \R^{\ptsNum \times 3}$ is then derived by back-projecting and downsampling the segmented depth image, where $\ptsNum$ denotes the number of points. Taking $\img$ and $\pts$ as inputs, our \ours{} predicts the 3D rotation $\rotation \in \so$, 3D translation $\translation \in \R^3$, and 3D size $\size \in \R^3$ of the observed object instance within predefined categories.

\begin{figure*}[t]
\centering
\includegraphics[width=1\linewidth]{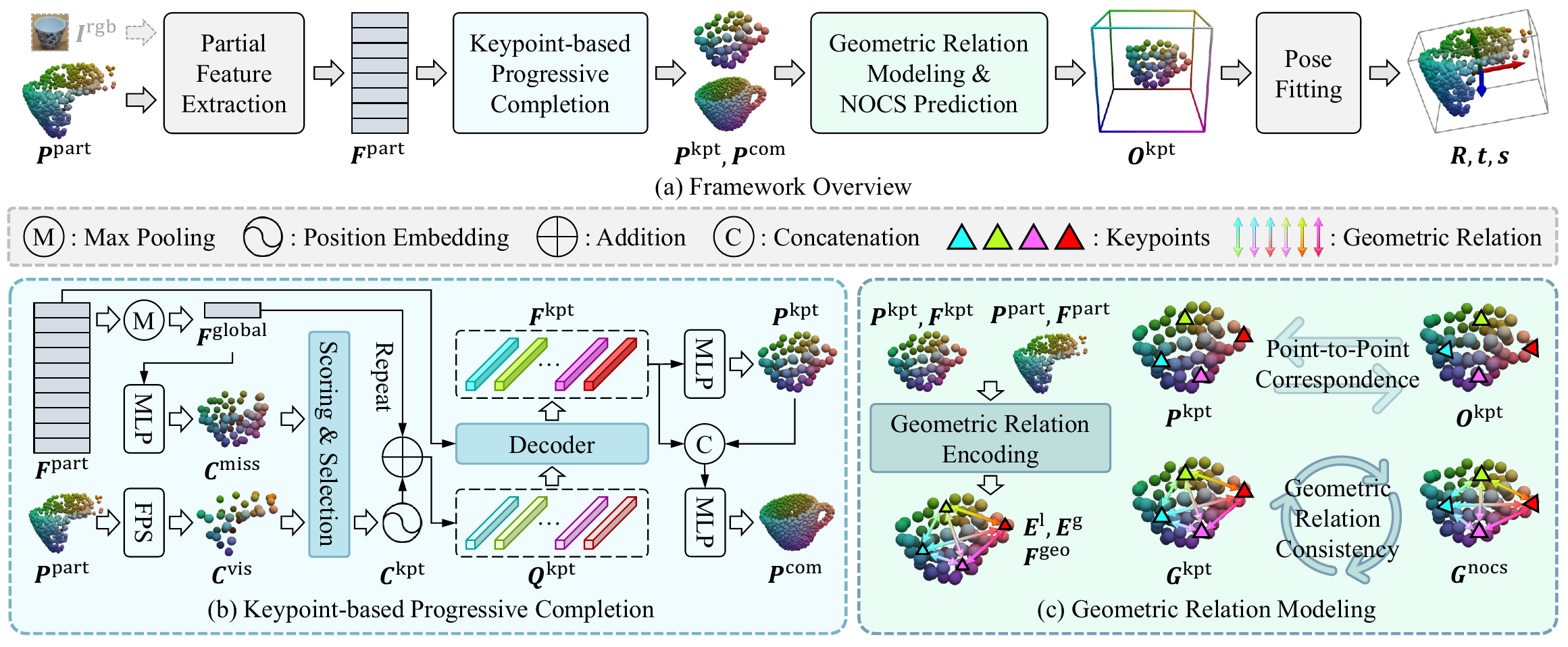}
\vspace{-1.75em}
\caption{%
(a) Overview of the proposed ComPose framework, which supports both RGB-D and depth-only settings, where the latter omits the RGB images $\img$. (b) The initial coarse keypoints $\coor^\textKpt$ are adaptively selected from missing and visible candidates $\{ \coor^\textMiss, \coor^\textVis \}$. These coarse keypoints are then progressively refined through feature interactions with the partial features $\ptsFeat$ to recover complete object geometries, including refined keypoints $\kpt$ and dense shapes $\com$. (c) The keypoint features $\kptFeat$ are enhanced into $\feat^\textGeo$ via geometric relation encoding, incorporating both local and global geometric context $\{ \geoLocal, \geoGlobal \}$. To ensure robust coordinate transformations, the pairwise geometric relations $\relation$ among keypoints are constrained to maintain alignment between the observation and canonical spaces.
}
\vspace{-0.75em}
\label{fig:method}
\end{figure*}

\parhead{Overview.}
As illustrated in \Figref{fig:method}, the proposed framework consists of four components: partial feature extraction (\Secref{sec:method_feature}), keypoint-based progressive completion (\Secref{sec:method_completion}), geometric relation encoding (\Secref{sec:method_geometric}), and correspondence-based pose estimation (\Secref{sec:method_pose}).

\subsection{Partial Feature Extraction}
\label{sec:method_feature}

For the partial point cloud $\pts$, we adopt PointNet++~\cite{pointnet++} to extract point-wise geometric features $\feat^\textPn$, which serve as the initial representation $\feat^\textInit \in \R^{\ptsNum \times \featDim}$. Under the RGB-D setting, we further follow SecondPose~\cite{cvpr2024secondpose} by employing DINOv2~\cite{dinov2} to extract pose-consistent semantic features $\feat^\textDino$ from the RGB image $\img$. These semantic features are associated~\cite{cvpr2025spotpose} with the 3D points in $\pts$, concatenated with the geometric features $\feat^\textPn$, and then projected into a $\featDim$-dimensional space to form the initial feature representation $\feat^\textInit \in \R^{\ptsNum \times \featDim}$. To better capture global context, a stack of Self-Attention (SA)~\cite{attention} layers are applied to $\feat^\textInit$, enabling dynamic interactions among all points and yielding the refined partial representation $\ptsFeat \in \R^{\ptsNum \times \featDim}$, which is formulated as below:
\begin{equation}
    \ptsFeat = \operatorname{SA}(\feat^\textInit + \operatorname{PE}(\pts)),
\end{equation}
\begin{equation}
     \operatorname{SA}(\mQ) = \Softmax ((\mQ \mW^Q)(\mQ \mW^K)^\top/\sqrt{D})(\mQ \mW^V),
\end{equation}
where $\mW^{Q/K/V} \in \R^{\featDim \times \featDim}$ are projection matrices, $\Softmax$ and $\top$ denote the Softmax and transpose operations, respectively. $\operatorname{PE}$ represents the learnable position embedding~\cite{attention}, which provides explicit spatial guidance for feature interactions.

\subsection{Keypoint-based Progressive Completion}
\label{sec:method_completion}

After extracting features from partial observations, we draw inspiration from \cite{tpami2023adapointr} and design a keypoint-based progressive completion module tailored for object pose estimation. Unlike typical point cloud completion approaches~\cite{cvpr2018foldingnet,3dv2018pcn} that assume canonical object alignments, our method handles the more challenging scenario where partial shapes are observed under arbitrary poses. The progressive process begins by generating an initial set of coarse keypoints $\coor^\textKpt \in \R^{\kptNum \times 3}$, where $\kptNum$ represents the number of keypoints. These keypoints are then progressively refined through feature interactions to recover complete object geometries, including both sparse keypoints $\kpt \in \R^{\kptNum \times 3}$ and dense shapes $\com \in \R^{\comNum \times 3}$ with $\comNum$ points. This module enables a more comprehensive understanding of object shapes, which is crucial for robust pose reasoning.

\parhead{Coarse Keypoint Generation.}
Given the partial features $\ptsFeat$, we first apply global max pooling to obtain a global representation $\vf^\textGlobal \in \R^\featDim$, followed by an MLP to predict a set of coarse keypoints $\coor^\textMiss \in \R^{\coorNum^\textMiss \times 3}$ that indicate potentially missing regions. Meanwhile, Farthest Point Sampling (FPS) is applied to the partial point cloud $\pts$ to acquire a set of visible keypoints $\coor^\textVis \in \R^{\coorNum^\textVis \times 3}$, providing reliable geometric cues from the visible regions. These two sets of keypoints contribute to the construction of the initial coarse keypoints $\coor^\textKpt$. Nevertheless, due to varying levels of incompleteness across different observations, relying on a fixed ratio of missing and visible keypoints lacks flexibility.
To adaptively select representative keypoints, we concatenate $\coor^\textMiss$ and $\coor^\textVis$ to form the candidate set $\coor^\textCand \in \R^{\coorNum^\textCand \times 3}$, which is then fed into a scoring MLP to predict scores $\score \in \R^{\coorNum^\textCand}$ for each candidate. The top $\kptNum$ keypoints are retained as $\coor^\textKpt$ after ranking, enabling a flexible balance between missing and visible regions.

\parhead{Progressive Shape Completion.}
To further refine the initially generated coarse keypoints $\coor^\textKpt$ and achieve higher fidelity in shape recovery, we employ a Transformer~\cite{attention} decoder that facilitates fine-grained feature interactions with observations. Specifically, we first construct the keypoint queries $\mQ^\textKpt \in \R^{\kptNum \times \featDim}$ by fusing the position embedding of $\coor^\textKpt$ with the global feature $\vf^\textGlobal$, written as:
\begin{equation}
    \kptQuery = \operatorname{Repeat}(\vf^\textGlobal) + \operatorname{PE}(\coor^\textKpt).
\end{equation}
The position embedding of $\coor^\textKpt$ provides explicit spatial guidance for the subsequent feature interactions with the partial feature $\ptsFeat$ through a decoder composed of Cross-Attention (CA) and Self-Attention (SA) layers, yielding the refined keypoint features $\kptFeat \in \R^{\kptNum \times D}$, formulated as:
\begin{equation}
    \hat{\feat}^\textKpt = \operatorname{CA}(\kptQuery, \ptsFeat), \quad \kptFeat = \operatorname{SA}(\hat{\feat}^\textKpt),
\end{equation}
\begin{equation}\hspace{-1mm}
    \operatorname{CA}(\mQ, \mK) = \Softmax ((\mQ \mW^Q)(\mK \mW^K)^\top/\sqrt{D})(\mK \mW^V).
\end{equation}
These features are then passed through an MLP to directly predict the 3D keypoint coordinates $\kpt$. To reconstruct a dense and complete point cloud $\com$, each keypoint feature $\kptFeat_\kptInd$ is concatenated with its associated coordinate $\kpt_\kptInd$ and passed through an MLP, whose output is reshaped to form
the fine-grained local geometry $\fold_\kptInd \in \R^{\foldNum \times 3}$ around each keypoint. The final complete dense point cloud $\com \in \R^{\comNum \times 3}$, consisting of $\comNum = \kptNum \foldNum$ points, is formed by aggregating all $\fold_\kptInd$ outputs.

\parhead{Supervision and Loss Functions.}
During training, to supervise the learning of progressive completion, the CAD model $\cad$ of the object is first transformed into the observation space using the ground-truth pose parameters $\{ \rotation^\textGt, \translation^\textGt, \size^\textGt \}$, resulting in the transformed model $\cadObs$:
\begin{equation}
    \cadObs = \| \size^\textGt \|_2 \rotation^\textGt \cad + \translation^\textGt.
\end{equation}
The progressive completion loss is then computed using the Chamfer Distance~\cite{chamfer} between the transformed model $\cadObs$ and the set $\comSet = \{ \coor^\textMiss, \kpt, \com \}$, defined as:
\begin{equation}
    \Ls^\textCom = \sum\nolimits_{\ptsSup \in \comSet} \operatorname{CD}(\ptsSup, \cadObs),
\end{equation}
\begin{equation}
\begin{aligned}
    \operatorname{CD}(\mX, \mY) = \frac{1}{|\mX|} &\sum\nolimits_{\vx \in \mX} \min_{\vy \in \mY} \| \vx - \vy \|_2^2\\
    + \frac{1}{|\mY|} &\sum\nolimits_{\vy \in \mY} \min_{\vx \in \mX} \| \vx - \vy \|_2^2.
\end{aligned}
\label{eq:cd}
\end{equation}
During the selection of representative keypoints, we aim to retain those that are close to the transformed model $\cadObs$ while filtering out outlier predictions in $\coor^\textMiss$ and outlier points in $\coor^\textVis$~\cite{cvpr2025spotpose}. To achieve this, we employ an MSE loss function on the predicted keypoint scores $\score$, defined as:
\begin{equation}
    \Ls^\textScore = \frac{1}{\coorNum^\textCand} \sum\nolimits_\kptInd (\score_\kptInd - \score^\textGt_\kptInd)^2,
\end{equation}
\begin{equation}
    \score^\textGt_\kptInd = \exp(-\dis_\kptInd / \temperature), \quad \dis_\kptInd = \min_{\vy \in \cadObs} \| \coor^\textCand_\kptInd - \vy \|_2,
\end{equation}
where $\temperature = 0.05$ denotes the temperature hyper-parameter.

\subsection{Geometric Relation Encoding}
\label{sec:method_geometric}

To further enhance the geometric contextual modeling of keypoint features, we follow AG-Pose~\cite{cvpr2024agpose} to explicitly encode their surrounding geometric relations. Specifically, for the $\kptInd$-th keypoint $\kpt_\kptInd$, its $\knnNum$ nearest neighbors $\knn_\kptInd \in \R^{\knnNum \times 3}$ are first identified from $\pts$, and their corresponding features $\knnFeat_\kptInd \in \R^{\knnNum \times \featDim}$ are retrieved from $\ptsFeat$. We then compute both local and global geometric relation embeddings, denoted as $\geoLocal_\kptInd \in \R^{\knnNum \times \featDim}$ and $\geoGlobal_\kptInd \in \R^{\kptNum \times \featDim}$, respectively, as formulated below:
\begin{equation}
    \geoLocal_\kptInd = \operatorname{MLP}( \operatorname{Repeat}(\kpt_\kptInd) - \knn_\kptInd),
\end{equation}
\begin{equation}
    \geoGlobal_\kptInd = \operatorname{MLP}( \operatorname{Repeat}(\kpt_\kptInd) - \kpt).
\end{equation}
Subsequently, the keypoint features $\kptFeat$ are progressively enriched with local and global geometric context through the following alternating enhancement process, yielding the geometric-aware keypoint features $\geoFeat \in \R^{\kptNum \times \featDim}$:
\begin{equation}
    \hat{\feat}^\textKpt_\kptInd = \operatorname{CA}( \kptFeat_\kptInd, \operatorname{MLP}(\knnFeat_\kptInd + \geoLocal_\kptInd) ),
\end{equation}
\begin{equation}
\begin{aligned}
    \geoFeat_\kptInd = \operatorname{MLP}( &\hat{\feat}^\textKpt_\kptInd + \operatorname{AvgPool}(\hat{\feat}^\textKpt)\\
    + &\operatorname{PE}(\kpt_\kptInd) + \operatorname{AvgPool}(\geoGlobal_\kptInd) ).
\end{aligned}
\end{equation}

\subsection{Correspondence-based Pose Estimation}
\label{sec:method_pose}

We adopt a correspondence-based paradigm~\cite{cvpr2025spotpose} for object pose estimation. Specifically, given the geometry-enhanced keypoint features $\geoFeat$, an MLP is employed to predict the corresponding NOCS coordinates $\nocs^\textKpt \in \R^{\kptNum \times 3}$. The object pose $\{ \rotation, \translation, \size \}$ is then solved from the correspondences between the keypoint coordinates $\kpt$ and their NOCS counterparts $\nocs^\textKpt$ with a pose fitting algorithm such as the Umeyama algorithm~\cite{umeyama} or a deep estimator~\cite{eccv2022dpdn}.

\parhead{Point-to-Point Correspondence Supervision.}
To supervise the learning of NOCS coordinates, previous works~\cite{cvpr2024agpose,cvpr2025spotpose,cvpr2025structure} commonly employ a straightforward point-wise coordinate regression loss, which directly penalizes the deviation between the predicted and ground-truth coordinates in a point-to-point manner. Specifically, the ground-truth NOCS coordinates $\nocs^\textGt$ are derived by applying the ground-truth pose parameters $\{ \rotation^\textGt, \translation^\textGt, \size^\textGt \}$ to the keypoint coordinates $\kpt$, which is formulated as below:
\begin{equation}
    \nocs^\textGt = \frac{1}{\| \size^\textGt \|_2} (\rotation^\textGt)^\top (\kpt - \translation^\textGt).
\end{equation}
The correspondence loss is then defined as the point-wise $\normltwo$ or Smooth-$\normlone$~\cite{eccv2022dpdn} distance between the predicted NOCS coordinates $\nocs^\textKpt$ and the ground-truth $\nocs^\textGt$. For instance, the $\normltwo$-based loss is defined as:
\begin{equation}
    \Ls^\textCorr = \frac{1}{\kptNum} \sum\nolimits_{\kptInd} \| \nocs^\textKpt_\kptInd - \nocs^\textGt_\kptInd \|_2.
\end{equation}

\parhead{Geometric Relation Consistency Constraint.}
Nevertheless, the above point-to-point constraint fails to capture the holistic geometric structure of object. For example, two sets of NOCS coordinates may exhibit similar mean point-wise errors, yet represent substantially different overall shapes. This ambiguity can lead to a structural mismatch between the predicted NOCS coordinates and the underlying object geometry observed in the input space. As a result, it becomes challenging to reliably estimate a globally coherent rigid transformation from these correspondences, which may compromise the accuracy of the final pose estimation. To address this limitation, we propose a novel geometric relation consistency loss that explicitly enforces alignment in the overall geometric relations among keypoints to capture higher-order structural cues. Specifically, we compute the pairwise $\normltwo$ distances among the scaled keypoint coordinates $\kpt / \| \size^\textGt \|_2$ to construct the reference geometric relation matrix $\relation^\textKpt \in \R^{\kptNum \times \kptNum}$, and similarly derive the predicted counterpart $\relation^\textNocs$ from the predicted NOCS coordinates $\nocs^\textKpt$.
The geometric relation consistency loss is then computed between these two matrices, defined as:
\begin{equation}
    \Ls^\textGeo = \frac{1}{\kptNum \times \kptNum} \sum\nolimits_{\kptInd, \kptIndSecond} (\relation^\textKpt_{\kptInd, \kptIndSecond} - \relation^\textNocs_{\kptInd, \kptIndSecond})^2.
\end{equation}

In summary, the overall loss function is as follows:
\begin{small}
\begin{equation}
    \Ls^\textAll = \lambda^\textCom \Ls^\textCom + \lambda^\textScore \Ls^\textScore + \lambda^\textCorr \Ls^\textCorr + \lambda^\textGeo \Ls^\textGeo,
\end{equation}
\end{small}
with $\lambda^\textCom, \lambda^\textScore, \lambda^\textCorr, \lambda^\textGeo$ as balance hyper-parameters.

\section{Experiments}
\label{sec:experiments}

\subsection{Experimental Setup}
\label{sec:experimental_setup}

\parhead{Datasets.}
We conduct experiments on three benchmarks including CAMERA25, REAL275~\cite{cvpr2019nocs} and HouseCat6D~\cite{cvpr2024housecat6d}. CAMERA25 is a synthetic dataset comprising 275K training and 25K testing images across 6 object categories. These images are generated via mixed-reality techniques by rendering foreground objects onto real backgrounds. REAL275 is a real-world dataset with 4.3K training images from 7 scenes and 2.75K testing images from 6 scenes, covering the same 6 object categories as CAMERA25. HouseCat6D is an emerging real-world benchmark containing 20K training images from 34 scenes and 3K testing images from 5 scenes, spanning 10 household object categories. This dataset includes photometrically challenging objects with diverse viewpoints and \textit{occlusions}, posing significant challenges for accurate object pose estimation.

\parhead{Evaluation Metrics.}
Following previous works~\citep{cvpr2024agpose,cvpr2025spotpose}, we report the mean Average Precision (mAP) of $n^{\circ}m$ cm for 6D pose evaluation, which measures the percentage of predictions with rotation error below $n^{\circ}$ and translation error below $m$ cm. We also report the mAP of 3D Intersection over Union (\iou{x}) at a threshold of $x\%$ for joint 6D pose and 3D size evaluation.

\parhead{Implementation Details.}
For a fair comparison, we use the same instance segmentation masks as AG-Pose~\cite{cvpr2024agpose}, obtained from Mask R-CNN~\cite{maskrcnn}. We sample $\ptsNum = 1024$ partial points and extract $\kptNum = 64$ keypoints. The dense complete shape is reconstructed with $\comNum = 1024$ points, where each keypoint is expanded into $\foldNum = 16$ fine-grained local geometry points. The numbers of missing and visible keypoints are set to $\coorNum^\textMiss = 64$ and $\coorNum^\textVis = 32$, respectively. The network architecture includes 2 attention layers in each of the following modules: partial feature extraction, progressive shape completion, and geometric relation encoding. In the geometric relation encoding module, the number of nearest neighbors $\knnNum$ is set to 16 and 32 across the two layers, respectively. The feature dimension $\featDim$ is set to 256 for RGB-D input and 128 for the depth-only setting. For loss balancing, we set the weights $\lambda^\textCom = 15$, $\lambda^\textScore = 1$, $\lambda^\textCorr = 2$, and $\lambda^\textGeo = 1$. The network is trained using the Adam~\cite{adam} optimizer with an initial learning rate of 0.001 and a cosine annealing schedule. All experiments are conducted on a single RTX3090Ti GPU with a batch size of 24 over 200K iterations.

\subsection{Comparison with State-of-the-art Methods}

\begin{table}[t]
\centering
\setlength{\tabcolsep}{1mm}
\caption{%
Performance comparison with state-of-the-art methods on the REAL275 dataset. The method marked with `*' is reproduced by us. ``Prior'' refers to shape priors~\cite{eccv2020spd}. For each data setting, the best results are in \textbf{bold}, and the second best results are \underline{underlined}.
}
\label{tab:sota_real}
\vspace{-0.5em}
\resizebox{\linewidth}{!}{%
\begin{tabular}{lccccccc}
\toprule
\multicolumn{1}{l|}{Method}                               & \multicolumn{1}{c|}{Prior} & \iou{50}      & \multicolumn{1}{c|}{\iou{75}}      & \ducm{5}{2}   & \ducm{5}{5}   & \ducm{10}{2}  & \ducm{10}{5}  \\ \midrule \midrule
\multicolumn{8}{c}{\cellcolor[HTML]{F2F2F2}\textit{RGB-D Setting}}                                                                                                                                                \\ \midrule
\multicolumn{1}{l|}{SPD~\cite{eccv2020spd}}               & \multicolumn{1}{c|}{\checkmark}  & 77.3          & \multicolumn{1}{c|}{53.2}          & 19.3          & 21.4          & 43.2          & 54.1          \\
\multicolumn{1}{l|}{SGPA~\cite{iccv2021sgpa}}             & \multicolumn{1}{c|}{\checkmark}  & 80.1          & \multicolumn{1}{c|}{61.9}          & 35.9          & 39.6          & 61.3          & 70.7          \\
\multicolumn{1}{l|}{DPDN~\cite{eccv2022dpdn}}             & \multicolumn{1}{c|}{\checkmark}  & 83.4          & \multicolumn{1}{c|}{76.0}          & 46.0          & 50.7          & 70.4          & 78.4          \\
\multicolumn{1}{l|}{GCE-Pose~\cite{cvpr2025gcepose}}      & \multicolumn{1}{c|}{\checkmark}  & \textbf{84.1} & \multicolumn{1}{c|}{79.8}          & 57.0          & 65.1          & 75.6          & 86.3          \\
\multicolumn{1}{l|}{VI-Net~\cite{iccv2023vinet}}          & \multicolumn{1}{c|}{$\times$}    & -             & \multicolumn{1}{c|}{-}             & 50.0          & 57.6          & 70.8          & 82.1          \\
\multicolumn{1}{l|}{SecondPose~\cite{cvpr2024secondpose}} & \multicolumn{1}{c|}{$\times$}    & -             & \multicolumn{1}{c|}{-}             & 56.2          & 63.6          & 74.7          & 86.0          \\
\multicolumn{1}{l|}{AG-Pose~\cite{cvpr2024agpose}}        & \multicolumn{1}{c|}{$\times$}    & \textbf{84.1} & \multicolumn{1}{c|}{80.1}          & 57.0          & 64.6          & 75.1          & 84.7          \\
\multicolumn{1}{l|}{SpherePose~\cite{iclr2025spherepose}} & \multicolumn{1}{c|}{$\times$}    & \underline{84.0}    & \multicolumn{1}{c|}{79.0}          & 58.2          & \underline{67.5}    & 76.2          & \underline{88.2}    \\
\multicolumn{1}{l|}{SpotPose~\cite{cvpr2025spotpose}}     & \multicolumn{1}{c|}{$\times$}    & \textbf{84.1} & \multicolumn{1}{c|}{\underline{81.2}} & 59.7          & 64.8          & \underline{81.5}    & \underline{88.2}    \\
\multicolumn{1}{l|}{CleanPose~\cite{iccv2025cleanpose}}   & \multicolumn{1}{c|}{$\times$}    &         -     & \multicolumn{1}{c|}{-}             & \underline{61.5}    & 67.4          & 78.3          & 86.2          \\ \midrule
\multicolumn{1}{l|}{\textbf{ComPose}}                     & \multicolumn{1}{c|}{$\times$}    & \underline{84.0}    & \multicolumn{1}{c|}{\textbf{81.4}} & \textbf{62.1} & \textbf{68.0} & \textbf{81.8} & \textbf{89.2} \\ \midrule \midrule
\multicolumn{8}{c}{\cellcolor[HTML]{F2F2F2}\textit{Depth-only Setting}}                                                                                                                                           \\ \midrule
\multicolumn{1}{l|}{SAR-Net~\cite{cvpr2022sarnet}}        & \multicolumn{1}{c|}{\checkmark}  & 79.3          & \multicolumn{1}{c|}{62.4}          & 31.6          & 42.3          & 50.3          & 68.3          \\
\multicolumn{1}{l|}{RBP-Pose~\cite{eccv2022rbppose}}      & \multicolumn{1}{c|}{\checkmark}  & -             & \multicolumn{1}{c|}{67.8}          & 38.2          & 48.1          & 63.1          & 79.2          \\
\multicolumn{1}{l|}{DR-Pose~\cite{iros2023drpose}}       & \multicolumn{1}{c|}{\checkmark}  & 78.9          & \multicolumn{1}{c|}{68.2}          & 41.7          & 46.0          & 67.7          & 76.3          \\
\multicolumn{1}{l|}{GPV-Pose~\cite{cvpr2022gpvpose}}      & \multicolumn{1}{c|}{$\times$}    & -             & \multicolumn{1}{c|}{64.4}          & 32.0          & 42.9          & -             & 73.3          \\
\multicolumn{1}{l|}{HS-Pose~\cite{cvpr2023hspose}}        & \multicolumn{1}{c|}{$\times$}    & 82.1          & \multicolumn{1}{c|}{74.7}          & 46.5          & 55.2          & 68.6          & 82.7          \\
\multicolumn{1}{l|}{Query6DoF~\cite{iccv2023query6dof}}   & \multicolumn{1}{c|}{$\times$}    & \underline{82.5}    & \multicolumn{1}{c|}{\underline{76.1}}    & \underline{49.0}    & \underline{58.9}    & \underline{68.7}    & \underline{83.0}    \\
\multicolumn{1}{l|}{AG-Pose*~\cite{cvpr2024agpose}}       & \multicolumn{1}{c|}{$\times$}    & \textbf{83.2} & \multicolumn{1}{c|}{75.6}          & 48.8          & 58.8          & 68.5          & 80.8          \\ \midrule
\multicolumn{1}{l|}{\textbf{ComPose}}                     & \multicolumn{1}{c|}{$\times$}    & 82.1          & \multicolumn{1}{c|}{\textbf{77.0}} & \textbf{55.6} & \textbf{61.3} & \textbf{77.8} & \textbf{85.0} \\ \bottomrule
\end{tabular}%
}
\vspace{-1em}
\end{table}

\parhead{Results on the REAL275 dataset.}
Table~\ref{tab:sota_real} presents a comprehensive comparison between our method and existing RGB-D and depth-only approaches on the REAL275 dataset. Under the depth-only setting, our \ours{} outperforms previous methods by a large margin across all 6D pose evaluation metrics. Notably, when compared to the keypoint-based AG-Pose~\cite{cvpr2024agpose}, our \ours{} achieves significant improvements of 6.8\% on \ducm{5}{2} and 9.3\% on \ducm{10}{2}, demonstrating the importance of shape completion for precise pose estimation. For 3D IoU metrics, \ours{} achieves comparable performance on \iou{50} and sets a new state-of-the-art on the stricter \iou{75} criterion. When incorporating semantic information from RGB images, \ours{} exhibits further performance gains under the RGB-D setting, achieving the best results across all 6D pose metrics without relying on shape priors. These results validate the effectiveness of our unified completion-pose framework.

\begin{table}[t]
\centering
\setlength{\tabcolsep}{1.7mm}
\caption{%
Performance comparison with state-of-the-art methods on the HouseCat6D dataset. The method marked with `*' is reproduced by us. For each data setting, the best results are shown in \textbf{bold}, and the second best results are \underline{underlined}.
}
\label{tab:sota_housecat}
\vspace{-0.5em}
\resizebox{\linewidth}{!}{%
\begin{tabular}{lcccccc}
\toprule
\multicolumn{1}{l|}{Method}                               & \iou{25}      & \multicolumn{1}{c|}{\iou{50}}      & \ducm{5}{2}   & \ducm{5}{5}   & \ducm{10}{2}  & \ducm{10}{5}  \\ \midrule \midrule
\multicolumn{7}{c}{\cellcolor[HTML]{F2F2F2}\textit{RGB-D Setting}}                                                                                                             \\ \midrule
\multicolumn{1}{l|}{VI-Net~\cite{iccv2023vinet}}          & 80.7          & \multicolumn{1}{c|}{56.4}          & 8.4           & 10.3          & 20.5          & 29.1          \\
\multicolumn{1}{l|}{SecondPose~\cite{cvpr2024secondpose}} & 83.7          & \multicolumn{1}{c|}{66.1}          & 11.0          & 13.4          & 25.3          & 35.7          \\
\multicolumn{1}{l|}{AG-Pose~\cite{cvpr2024agpose}}        & 88.1          & \multicolumn{1}{c|}{76.9}          & 21.3          & 22.1          & 51.3          & 54.3          \\
\multicolumn{1}{l|}{SpherePose~\cite{iclr2025spherepose}} & 88.8          & \multicolumn{1}{c|}{72.2}          & 19.3          & \underline{25.9}    & 40.9          & 55.3          \\
\multicolumn{1}{l|}{SpotPose~\cite{cvpr2025spotpose}}     & 89.1          & \multicolumn{1}{c|}{77.0}          & 23.8          & 24.5          & 52.3          & 54.8          \\
\multicolumn{1}{l|}{GCE-Pose~\cite{cvpr2025gcepose}}      & -             & \multicolumn{1}{c|}{79.2}          & \underline{24.8}    & 25.7          & \underline{55.4}    & \underline{58.4}    \\
\multicolumn{1}{l|}{CleanPose~\cite{iccv2025cleanpose}}   & \underline{89.2}    & \multicolumn{1}{c|}{\underline{79.8}}    & 22.4          & 24.1          & 51.6          & 56.5          \\ \midrule
\multicolumn{1}{l|}{\textbf{ComPose}}                     & \textbf{90.3} & \multicolumn{1}{c|}{\textbf{80.6}} & \textbf{25.8} & \textbf{27.6} & \textbf{57.8} & \textbf{61.5} \\ \midrule \midrule
\multicolumn{7}{c}{\cellcolor[HTML]{F2F2F2}\textit{Depth-only Setting}}                                                                                                        \\ \midrule
\multicolumn{1}{l|}{FS-Net~\cite{cvpr2021fsnet}}          & 74.9          & \multicolumn{1}{c|}{48.0}          & 3.3           & 4.2           & 17.1          & 21.6          \\
\multicolumn{1}{l|}{GPV-Pose~\cite{cvpr2022gpvpose}}      & 74.9          & \multicolumn{1}{c|}{50.7}          & 3.5           & 4.6           & 17.8          & 22.7          \\
\multicolumn{1}{l|}{AG-Pose*~\cite{cvpr2024agpose}}       & \underline{81.4}    & \multicolumn{1}{c|}{\underline{59.9}}    & \underline{9.7}     & \underline{10.6}    & \underline{25.9}    & \underline{29.7}    \\ \midrule
\multicolumn{1}{l|}{\textbf{ComPose}}                     & \textbf{81.6} & \multicolumn{1}{c|}{\textbf{65.1}} & \textbf{11.8} & \textbf{12.7} & \textbf{34.8} & \textbf{38.9} \\ \bottomrule
\end{tabular}%
}
\vspace{-1em}
\end{table}

\parhead{Results on the HouseCat6D dataset.}
Table~\ref{tab:sota_housecat} reports the performance comparison on the more challenging HouseCat6D dataset. Our ComPose consistently achieves state-of-the-art results across all evaluation metrics under both depth-only and RGB-D settings. Specifically, under the depth-only setting, ComPose outperforms AG-Pose~\cite{cvpr2024agpose} by 5.2\% on the \iou{50} metric and 2.1\% on the \ducm{5}{2} metric. Under the RGB-D setting, ComPose surpasses GCE-Pose~\cite{cvpr2025gcepose} by 1.4\% on \iou{50} and 1.0\% on \ducm{5}{2}. The superior performance on this more comprehensive and challenging real-world dataset with occlusions further validate the effectiveness and robustness of our \ours{} framework.

\parhead{Results of Completion.}
Previous methods typically rely on category-level shape priors~\cite{eccv2020spd} to reconstruct object CAD models at unit scale \textit{in the canonical space}. In contrast, our approach is the first to perform shape completion directly \textit{in the observation space} under arbitrary poses, which is much more challenging. \Cref{tab:sota_cd_distance_camera} compares the reconstruction performance of different methods on the hard camera category of the REAL275 dataset. To ensure a fair comparison, in addition to the metric-scale CD as defined in \Cref{eq:cd}, we also compute the unit-scale CD$^{\text{unit}}$ metric. This involves normalizing the completed shape $\com$ and the ground-truth shape $\cadObs$ to unit scale using the ground-truth scale $\|\size^\textGt\|_2$ before calculating the Chamfer Distance. As shown in the table, without relying on shape priors, our RGB-D model performs best in reconstructing complete 3D shapes. 

\begin{table}[t]
\centering
\setlength{\tabcolsep}{3.2mm}
\caption{%
Reconstruction performance comparisons for the camera category on the REAL275 dataset, measured using Chamfer Distance ($\times 10^{-3}$). CD$^{\text{unit}}$ is computed after \textit{unit-scale} normalization of the shape, while CD is computed directly in the observation space under the real-world \textit{metric scale}. A lower value ($\downarrow$) indicates better performance, with the best results highlighted in \textbf{bold}.
}
\label{tab:sota_cd_distance_camera}
\vspace{-0.75em}
\resizebox{\linewidth}{!}{%
\begin{tabular}{lcccc}
\toprule
\multicolumn{1}{l|}{Method}                        & \multicolumn{1}{c|}{Data Setting} & \multicolumn{1}{c|}{Shape Prior} & CD$^{\text{unit}}$ $\downarrow$ & CD $\downarrow$ \\ \midrule \midrule
\multicolumn{5}{c}{\cellcolor[HTML]{F2F2F2}\textit{Reconstruct 3D Object Models in the Canonical Space}}                                                                        \\ \midrule
\multicolumn{1}{l|}{SPD~\cite{eccv2020spd}}        & \multicolumn{1}{c|}{RGB-D}        & \multicolumn{1}{c|}{\checkmark}  & 8.89            & -                                 \\
\multicolumn{1}{l|}{SGPA~\cite{iccv2021sgpa}}      & \multicolumn{1}{c|}{RGB-D}        & \multicolumn{1}{c|}{\checkmark}  & 5.51            & -                                 \\
\multicolumn{1}{l|}{DR-Pose~\cite{iros2023drpose}} & \multicolumn{1}{c|}{D}            & \multicolumn{1}{c|}{\checkmark}  & 5.26            & -                                 \\ \midrule \midrule
\multicolumn{5}{c}{\cellcolor[HTML]{F2F2F2}\textit{Reconstruct Complete 3D Shapes in the Observation Space}}                                                                    \\ \midrule
\multicolumn{1}{l|}{\textbf{ComPose}}              & \multicolumn{1}{c|}{RGB-D}        & \multicolumn{1}{c|}{$\times$}    & \textbf{4.20}   & \textbf{0.17}                     \\
\multicolumn{1}{l|}{\textbf{ComPose}}              & \multicolumn{1}{c|}{D}            & \multicolumn{1}{c|}{$\times$}    & 6.09            & 0.23                              \\ \bottomrule
\end{tabular}%
}
\vspace{-0.75em}
\end{table}

\begin{table}[t]
\centering
\setlength{\tabcolsep}{1.4mm}
\caption{%
Performance comparison of different depth-only methods under occlusion-augmented testing on the REAL275 dataset.
}
\label{tab:sota_occlusion}
\vspace{-0.75em}
\resizebox{\linewidth}{!}{%
\begin{tabular}{l|c|cccc}
\toprule
Method (D)                                          & Test with OccAug  & \ducm{5}{2} & \ducm{5}{5} & \ducm{10}{2} & \ducm{10}{5} \\ \midrule
\multirow{3}{*}{AG-Pose*~\cite{cvpr2024agpose}} & $\times$          & 48.8        & 58.8        & 68.5         & 80.8         \\
                                                & \checkmark        & 37.1        & 49.1        & 54.3         & 72.6         \\
                                                & Drop $\downarrow$ & 24.0\%      & 16.5\%      & 20.7\%       & 10.1\%       \\ \midrule
\multirow{3}{*}{\textbf{ComPose}}               & $\times$          & 55.6        & 61.3        & 77.8         & 85.0         \\
                                                & \checkmark        & 42.7        & 53.6        & 62.9         & 77.7         \\
                                                & Drop $\downarrow$ & \textbf{23.2\%} & \textbf{12.6\%} & \textbf{19.2\%} & \textbf{8.6\%} \\ \bottomrule
\end{tabular}%
}
\vspace{-0.75em}
\end{table}

\parhead{Robustness to Occlusion.}
Although the integrated shape completion primarily focuses on addressing point cloud incompleteness caused by \textit{self-occlusion}, our method also performs effectively under \textit{external occlusion}. In \Cref{tab:sota_occlusion}, we simulate severe occlusion by applying a 25\% occlusion mask to the object segmentation masks on the REAL275 dataset, with the masked region randomly selected from the top, bottom, left, and right edges of the object.
As shown, AG-Pose suffers a performance drop of 16.5\% on \ducm{5}{5}, while our ComPose exhibits a smaller decrease by 12.6\%, demonstrating its superior robustness against occlusion.

\subsection{Ablation Studies}
\label{sec:ablation}

In this section, we conduct comprehensive ablation studies on the REAL275 dataset under the depth-only setting to shed more light on the superiority of our method.

\begin{table}[t]
\centering
\setlength{\tabcolsep}{1.4mm}
\caption{%
Ablation studies on the shape completion strategy.
``Partial Instance'' indicates reconstructing only visible object regions.
}
\label{tab:ablation_completion}
\vspace{-0.75em}
\resizebox{\linewidth}{!}{%
\begin{tabular}{l|cc|cccc}
\toprule
Reconstruction                & $\kpt$     & $\com$     & \ducm{5}{2}   & \ducm{5}{5}   & \ducm{10}{2}  & \ducm{10}{5}  \\ \midrule
Partial Instance~\cite{cvpr2024agpose} & \checkmark & \checkmark & 49.6          & 56.1          & 72.4          & 82.0          \\
\rowcolor[HTML]{F2F2F2} 
Complete $\cadObs$            & \checkmark & \checkmark & \textbf{55.6} & \textbf{61.3} & \textbf{77.8} & \textbf{85.0} \\
Complete $\cadObs$            & \checkmark & $\times$   & 54.9          & 60.5          & 76.1          & 83.3          \\ \bottomrule
\end{tabular}%
}
\vspace{-1em}
\end{table}

\parhead{Efficacy of Shape Completion.}
\Cref{tab:ablation_completion} presents ablation studies on different shape completion strategies. Replacing the \textit{complete shape recovery} with the \textit{partial instance reconstruction} as in AG-Pose~\cite{cvpr2024agpose} results in a 6\% decrease on \ducm{5}{2}, demonstrating the importance of complete geometric cues provided by shape completion for precise pose estimation.
Moreover, the completion of dense point clouds $\com$ enhances the fine-grained geometric awareness of keypoints, leading to a 1.7\% performance gain on \ducm{10}{5}.

\parhead{Efficacy of the Unified Framework.}
\Cref{fig:motivation_accuracy_fps} illustrates the superiority of the proposed unified Completion-Pose framework over two-stage Completion-then-Pose scheme in terms of both accuracy and efficiency. In the two-stage pipeline, the completed dense shape $\com$ replaces the original partial input $\pts$ of AG-Pose~\cite{cvpr2024agpose}, introducing additional inference time during completion. In contrast, ComPose directly replaces AG-Pose's \textit{keypoint detection module} with the \textit{keypoint-based completion module}, eliminating extra latency. Both frameworks utilize the same completion module, with the depth-only version operating at a feature dimension of 128. The efficiency gains of ComPose over the original AG-Pose mainly stem from reducing the keypoint count from 96 to 64, removing the Self-Attention operation in NOCS prediction, and using the Umeyama algorithm rather than a deep estimator for pose fitting.

\begin{table}[t]
\centering
\setlength{\tabcolsep}{0.9mm}
\caption{%
Ablation studies on the progressive completion process, where $\kptNum$ is kept constant at 64 across all experiments.
}
\label{tab:ablation_progressive}
\vspace{-0.75em}
\resizebox{\linewidth}{!}{%
\begin{tabular}{l|ccc|cccc}
\toprule
Completion                          & Selection  & $\coorNum^\textMiss$ & $\coorNum^\textVis$ & \ducm{5}{2}   & \ducm{5}{5}   & \ducm{10}{2}  & \ducm{10}{5}  \\ \midrule
Static Query                        & -          & -                    & -                   & 51.6          & 58.6          & 72.9          & 82.0          \\
PoinTr~\cite{iccv2021pointr}        & $\times$   & 64                   & 0                   & 52.7          & 59.6          & 74.3          & 82.7          \\
AdaPoinTr~\cite{tpami2023adapointr} & \checkmark & 64                   & 32                  & 53.4          & 59.8          & 75.6          & 83.8          \\ \midrule
\rowcolor[HTML]{F2F2F2} 
Progressive                         & \checkmark & 64                   & 32                  & \textbf{55.6} & \textbf{61.3} & \textbf{77.8} & \textbf{85.0} \\
Progressive                         & $\times$   & 64                   & 0                   & 53.8          & 61.0          & 75.4          & 84.0          \\
Progressive                         & $\times$   & 32                   & 32                  & 54.6          & 60.0          & 76.6          & 83.7          \\ \bottomrule
\end{tabular}%
}
\vspace{-0.85em}
\end{table}

\begin{table}[t]
\centering
\setlength{\tabcolsep}{2.7mm}
\caption{%
Ablation studies on the geometric relation modeling.
}
\label{tab:ablation_geo_modeling}
\vspace{-0.75em}
\resizebox{\linewidth}{!}{%
\begin{tabular}{cc|cccc}
\toprule
Encoding   & Consistency                & \ducm{5}{2}   & \ducm{5}{5}   & \ducm{10}{2}  & \ducm{10}{5}  \\ \midrule
$\times$   & $\times$                   & 49.5       & 55.6          & 71.9          & 79.7          \\
\checkmark & $\times$                   & 53.8          & 60.5          & 74.8          & 83.5          \\
\rowcolor[HTML]{F2F2F2} 
\checkmark & \checkmark                 & \textbf{55.6} & \textbf{61.3} & \textbf{77.8} & \textbf{85.0} \\ \bottomrule
\end{tabular}
}
\vspace{-1.25em}
\end{table}

\parhead{Efficacy of Keypoint-based Progressive Completion.}
In \Cref{tab:ablation_progressive}, we conduct ablation studies on the progressive completion process. ``Static Query'' indicates that keypoint queries $\kptQuery$ are initialized as learnable embeddings~\cite{tip2024apl}, without the construction of coarse keypoints $\coor^\textKpt$. Unlike our approach, classic methods such as PoinTr~\cite{iccv2021pointr} and AdaPoinTr~\cite{tpami2023adapointr} do not enforce constraints on the reconstruction of $\coor^\textMiss$, resulting in a significant deviation of the predicted coarse coordinates $\coor^\textKpt$ from the \textit{actual object shapes}. In addition, AdaPoinTr lacks explicit supervision during the keypoint selection process. As shown in the table, our progressive completion module outperforms all three alternative strategies, with a 2.2\% improvement over AdaPoinTr on the \ducm{5}{2} metric. The last two rows further demonstrate the necessity of adaptive keypoint selection, which effectively harnesses reliable geometric cues from visible keypoints and flexibly balances missing and visible regions.

\parhead{Efficacy of Geometric Relation Modeling.}
We conduct ablation studies on the components of geometric relation modeling in \Cref{tab:ablation_geo_modeling}. The geometric relation encoding module explicitly enhances the keypoint features by capturing geometric context, leading to a 4.3\% improvement on the \ducm{5}{2} metric. Additionally, the geometric relation consistency constraint enforces structural alignment between the predicted NOCS coordinates and the observed object geometry, further improving performance by 1.8\% on \ducm{5}{2}.

\subsection{Visualization}
\begin{figure}[t]
\centering
\includegraphics[width=0.99\linewidth]{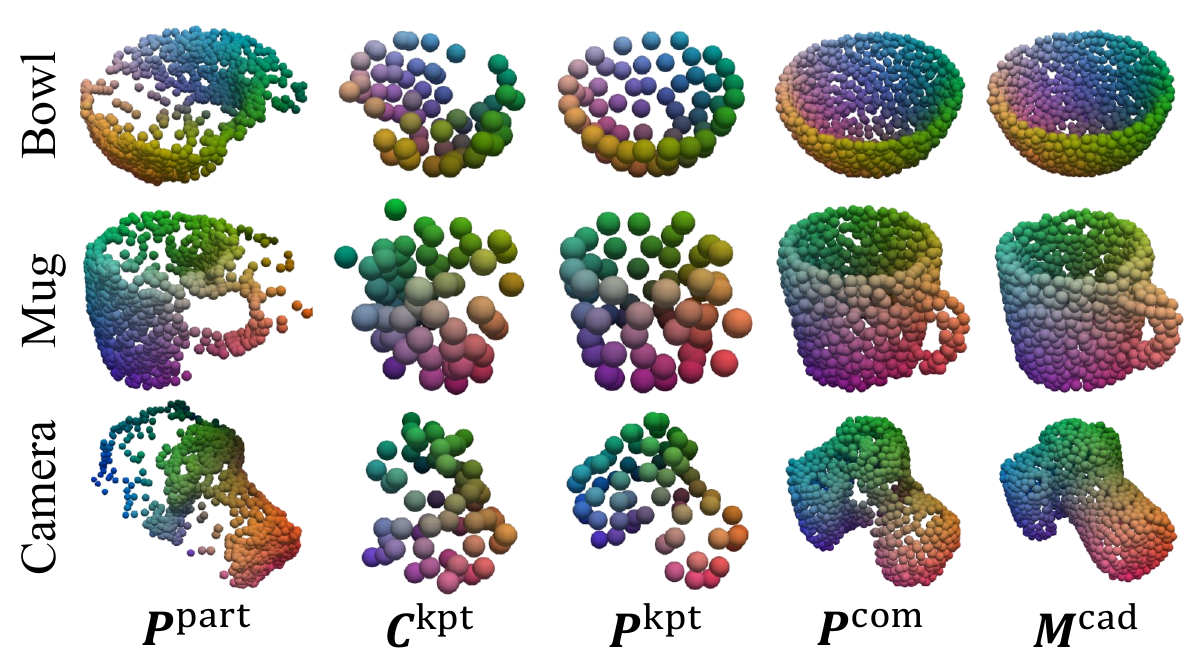}
\vspace{-0.75em}
\caption{%
Visualization of the keypoint-based progressive completion. Complete object geometries are progressively recovered.
}
\vspace{-0.75em}
\label{fig:visual_completion}
\end{figure}

\begin{figure}[t]
\centering
\includegraphics[width=0.99\linewidth]{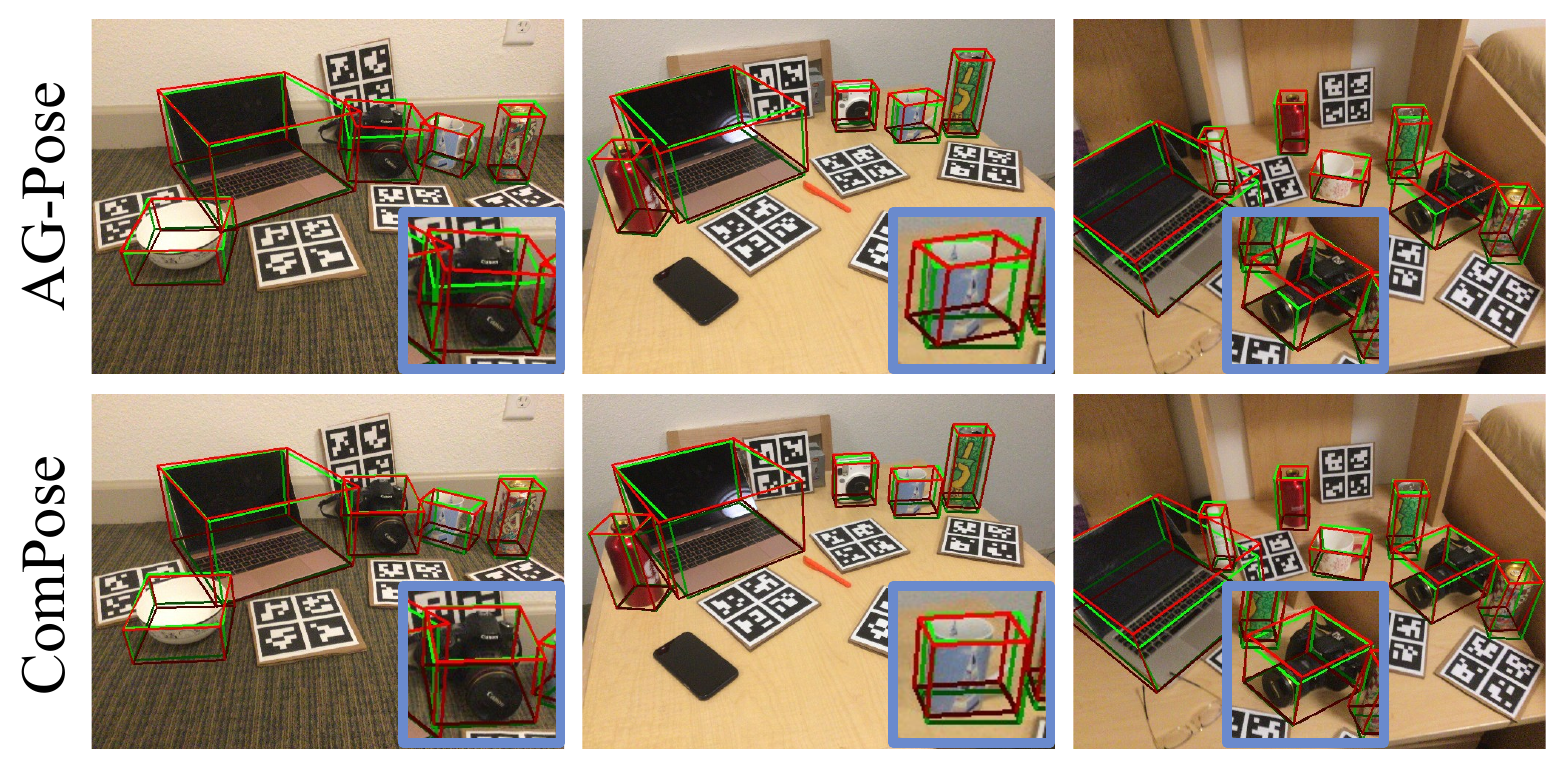}
\vspace{-0.75em}
\caption{%
Qualitative comparison between our \ours{} and AG-Pose~\cite{cvpr2024agpose}. Red/Green
indicates the predicted/GT results.
}
\vspace{-1em}
\label{fig:visual_compare}
\end{figure}

\Figref{fig:visual_completion} visualizes the keypoint-based progressive completion process. The results demonstrate that our completion module not only progressively recovers the complete object shape from partial observations, but also effectively filtering out outlier points caused by inaccurate segmentation~\cite{cvpr2025spotpose,pan2024purify,pan2024rethinking,pan2025exploring}, yielding a cleaner and more comprehensive object representation. This refined geometric representation enhances the robustness of object pose estimation. As presented in \Figref{fig:visual_compare}, our \ours{} achieves more accurate and reliable pose predictions than AG-Pose~\cite{cvpr2024agpose}, benefiting from a better comprehension of holistic object geometry.

\section{Conclusion}
\label{sec:conclusion}

In this work, we propose \ours{}, a novel framework that integrates keypoint-based progressive completion to enhance the understanding of complete object shapes, which exhibits significant potential for robust category-level object pose estimation. To further improve structural awareness, we introduce geometric relation encoding and consistency constraints, which explicitly capture object geometric structure and facilitate globally coherent coordinate transformations for robust pose fitting. Extensive experiments on existing benchmarks consistently demonstrate the superior performance of our method under both RGB-D and depth-only settings without relying on category-level shape priors.

\section*{Acknowledgement}

This work was supported by the National Natural Science Foundation of China (Grant No. 62306294) and the Open Fund of National Key Laboratory of Deep Space Exploration (Grant NKDSEL2025008).

\clearpage
{
    \small
    \bibliographystyle{ieeenat_fullname}
    \bibliography{main}
}


\end{document}